%% file: track_extraction_HRC.tex
\documentclass[final,twocolumn,10pt,letterpaper]{IEEEtran}
\usepackage{cuted}
\usepackage{lipsum, color}
\usepackage{amsmath}
\usepackage{amsthm}
\usepackage{subfigure}
\usepackage{verbatim}
\usepackage{amsfonts}
\usepackage{color}
\usepackage{graphicx}
\usepackage{epstopdf}
\usepackage{graphicx}
\usepackage{filecontents}
\usepackage[noadjust]{cite}
\newcommand{\beq}{\begin{equation}}
\newcommand{\eeq}{\end{equation}}

\newcommand{\Xc}{\mathcal{X}}
\newcommand{\Yc}{\mathcal{Y}}
\newcommand{\Sc}{\mathcal{S}}

\renewcommand{\Pr}{\mathbf{P}}
\newcommand{\Er}{\mathbf{E}}

\newcommand{\Rbb}{\mathbb{R}}

\newcommand{\lbr}{\left\{}
\newcommand{\rbr}{\right\}}
\newcommand{\lb}{\left(}
\newcommand{\rb}{\right)}

\newcommand{\bcr}{\begin{color}{red}}
\newcommand{\bcb}{\begin{color}{blue}}
\newcommand{\ec}{\end{color}}

\begin{document}

\title{Track Extraction with Hidden Reciprocal Chain Models}
\author{ George~Stamatescu,~\IEEEmembership{Student~Member,~IEEE,} Langford~B~White,~\IEEEmembership{Senior~Member,~IEEE,}  Riley Bruce-Doust*

 \thanks{George Stamatescu, Langford B White and Riley Bruce-Doust are with the School of
  Electrical and Electronic Engineering, The University of Adelaide, North Tce, Adelaide 5005, 
  Australia. \texttt{\{George.Stamatescu, Lang.White, Riley.Bruce-Doust\}@adelaide.edu.au}}

}

\markboth{DRAFT}{L.B. White}
\maketitle

\begin{abstract} 
 This paper develops Bayesian track extraction algorithms for targets modelled as hidden reciprocal chains (HRC). HRC are a class of finite-state random process models that generalise the familiar hidden Markov chains (HMC). HRC are able to model the ``intention'' of a target to proceed from a given origin to a destination, behaviour which cannot be properly captured by a HMC. While Bayesian estimation problems for HRC have previously been studied, this paper focusses principally on the problem of track extraction, of which the primary task is confirming target existence in a set of detections obtained from thresholding sensor measurements.  Simulation examples are presented which show that the additional model information contained in a HRC improves detection performance when compared to HMC models.

\end{abstract}


\section{Introduction} \label{sec:intro}
\input{intro_v8.tex} 

\section{Hidden Reciprocal Chain Models\label{sec:mod}}

In this section, we define a reciprocal chain (RC) and summarise the Markov bridge (MB) construction of a RC as outlined in \cite{WC11}. We then describe how source-destination awareness is encoded in the RC model, and how this awareness can be approximated with a Schr\"{o}dinger bridge. We then define a HRC as a natural generalisation of a RC in the same way that a HMC generalises a Markov chain (MC). The HRC construction is used to derive the optimal HRC filters in section \ref{sec:filt}. 

\subsection{The Markov Bridge Construction of a RC}

\input{MB_construction_v8.tex}

\subsection{Encoding Source Destination Awareness\label{ssec:encode_dest}}

As stated in the introduction, this paper compares the performance of a tracker derived from a reciprocal model to a tracker whose target model includes future information but remains within the Markovian class. Specifically, we will compare to the Schr\"{o}dinger bridge (SB) \cite{Jami75,PT,GP}. In this subsection we describe both the RC and SB approaches to encoding future information into a Markov process representing a target model with given Markov transition probability matrices, referred to as the base process. We then present an example of a non-Markov RC specified by $\Pi$.

\subsubsection*{Reciprocal chains from a base process}\

This construction was established in \cite{Jami74} and will be used here, as it was in \cite{WC11}, for numerical simulations in Section \ref{sec:sims}. Let $A$ denote the time-homogeneous transition probability matrix for a $N$ state Markov chain $Z_t$, which we call the base process, where the entries of the matrix are $A_{i,j} = \Pr \lbr Z_t = j | Z_t = i \rbr$. A reciprocal process can be constructed from the base process with three point transitions given by \cite{Jami74}
\begin{align*}
Q_{i,j,\ell}(t) = \frac{A_{i,j} A_{j,k}}{\sum_{\ell=1}^{N}A_{i,\ell} A_{\ell, k}} \ , 
 \end{align*}
for $t = 1, \ldots, T-2$. As described in \cite{WC11},  The corresponding MB transitions \eqref{eq:MBform2}, are now given by
\begin{align}
B_{i,j}^k(t) = \frac{A_{i,j} (A^{T-(t+1)})_{j,k}}{(A^{T-t})_{i,k}} \ ,
\ \label{eq:MB_As}
\end{align}
for $t = 0, \ldots, T-2$. As described in \cite{WC11}, the process for generating a sample path of this RC is to draw the initial and final points $X_0$ and $X_T$ from $\Pi$, which specifies the MB transitions corresponding to $X_T = k$. The sample path is then constructed in the standard way for a MC, starting from $X_0$ using the transitions \eqref{eq:MB_As}.

The end-points distribution obviously encodes the target's source-destination awareness, as well as determining whether a particular RC is Markov or not. An example of a non-Markov destination awareness which we call {\it loitering behaviour} where a target returns to its origin by some final time $T$, can be encoded via
\begin{align}
\Pr_{Loitering} \lbr X_0=i, X_T=j \rbr =
\begin{cases}
p_r(i), \quad \text{if } i=j, \\
0, \quad \text{otherwise}.
\label{eq:loitering_ends}
\end{cases}
\end{align}
where $p_r(i)$ is the probability of starting and returning to the origin i, and $\sum_{i=1}^{N} p_r(i) = 1$. If there is a uniform probability on loitering anywhere in a statespace with $N$ states, the end-points distribution for this loitering reciprocal chain (LRC) model can be written in matrix form;
\begin{eqnarray*}
\Pi_{LRC} &=&
 \begin{pmatrix}
  \frac{1}{N} & 0 & \cdots & 0 \\
 0&  \frac{1}{N} & \cdots & 0 \\
  \vdots  & \vdots  & \ddots & \vdots  \\
  0 & 0 & \cdots &  \frac{1}{N}
 \end{pmatrix}
\end{eqnarray*}\\
We can see clearly how ``source-destination pairs'' of a target are encoded via $\Pi$.

\subsubsection*{Schr\"{o}dinger bridges from a base process}\

The SB is the time-inhomogeneous process that attains a specified marginal distribution at its final time and which has transitions, i.e. `dynamics', closest to (in Kullback-Liebler information divergence sense) the specified {\it a priori} homogeneous dynamics of the base process (see \cite{GP} for a finite-state proof). We thus recount the Schr\"{o}dinger bridge transition matrix construction. The following originates from E. Schr\"{o}dinger's idea, was formalised in probabilistic terms by Jamison \cite{Jami75}, and was recently translated into the discrete-state setting by Pavon {\it et al} \cite{PT,GP}.
 Let $A$ denote the homogeneous state transition matrix of a base process $Z_t$, as before, and let  $X_t$ be the SB of $Z_t$ with marginal distributions $\pi_0$ and $\pi_T$ on $X_0$ and $X_T$ respectively. 
Let 
$\lambda_0, \lambda_T$ be the solutions of\begin{align}
\pi_T & = \lambda_T \circ \lambda_0A^{T} \nonumber \\
\pi_0 & = \lambda_0 \circ \lambda_TA'^{T} \nonumber  
\end{align} where $\circ$ is the element-wise product. (Existence and uniqueness is proved in \cite{GP}.)
Define the positive row vectors 
 $\psi_t =  \lambda_T  A'^{(T-t)}$.
 Then the transitions of the SB are given by
\beq
 S_{i,j}(t)= \Pr \lbr X_{t+1}= j| X_t = i\rbr =A_{i,j} \frac{\psi_{t+1}(j)}{\psi_t(i)}
 \label{eq:SB_trans}
\eeq

We can attempt to approximate a RC with a SB by setting the marginals $\pi_0$ and $\pi_T$ to be precisely those of the RC's endpoints distribution $\Pi$. As we will see in section \ref{sec:sims}, although SBs are a way to encode future information, they are unable to capture the specific source-destination awareness that non-Markov reciprocal processes can, as there cannot be an arbitrary statistical relationship between the start and end-points of a target. 

\subsection{Hidden Reciprocal Chains (HRCs)\label{ssec:HRC}}

\input{HRCs_ssec_v7.tex}


\section{Observation Models \label{sec:obs_mod}}

In this section, we define the two observation models which correspond to the tracking scenarios of interest, that is, a set of detections over an interval obtained by applying a threshold to sensor returns. The first model allows a single detection at each time, that may be a target detection or a false alarm, modelled with position uncertainty. The second model generalises this to allow multiple detections at each time, with at most one detection corresponding to a target. Note that the single observation model could alternatively be interpreted as potentially erroneous output from a fast-time tracker, such as a Kalman filter (see \cite{FK13} for further details), while the second model is more typical of general tracking scenarios, and similar to that of \cite{vK}.
\subsection{Single Observation Model} \label{ssec:single_obs}

\input{single_obs_mod_v8.tex}

\subsection{Multiple Observation Model} \label{ssec:multi_obs}

We can generalise the single observation case to $M$ multiple detections at a time $t$, with at most one target in the system and potentially no target detections at any particular time. Thus we have that the observations at each time $t$ are a list $Y_t = \lbr y^1_t, \dots y^M_t \rbr$.

We can define an \emph{association} random vector $\mathbf{a_t}$ which acts as a pointer to the list of observations, associating an observation to a target detection. It is convenient to let $\mathbf{a_t}$ take as values the zero vector or one of the orthonormal basis vectors in $\Rbb^M$, 
\begin{align*}
\mathbf{a_t} \in \lbr \mathbf{0, e_1, e_2, \dots, e_M} \rbr
\end{align*}
Hence $\mathbf{a_t}$ has a $1$ in at most one position, and zeros elsewhere, with entries corresponding to the entries of the list  $Y_t$. We could view each entry of $\mathbf{a_t}$ as a mode just as in the previous observation model. We denote the \emph{a priori} probability over $ \mathbf{a_t}$ as,
 \begin{align}
 \Pr \lbr \mathbf{a_t = 0}\rbr &= \lambda_0 \label{eq:multi_prior}\\
 \Pr \lbr \mathbf{a_t = e_{\ell}}\rbr &= \lambda_{\ell} \nonumber
 \end{align}
where $\lambda_0 \in [0,1]$, and clearly $\sum_{\ell = 1}^M \lambda_{\ell} =1-\lambda_0$
This means the observation vector likelihood $C_i(t) = \Pr \lbr Y_t | X_t = i \rbr$ can be expressed as the sum over disjoint events, with each event corresponding to an assignment of a particular observation $y^{\ell}_t$ to the target in some state, $X_t = i$, and all other observations false alarms from the sensor detector,  
\begin{align}
C_i(t) &= \Pr \lbr {Y}_t | X_t = i\rbr \nonumber \\
 &= \sum_{\ell = 1}^M \Pr \lbr{Y}_t, \mathbf{a_t = e_{\ell}}| X_t = i\rbr +\Pr \lbr {Y}_t, \mathbf{a_t = 0}| X_t = i\rbr \nonumber\\
  &= \sum_{\ell = 1}^M \Pr \lbr \mathbf{a_t = e_{\ell} }\rbr  \Pr \lbr{Y}_t | X_t = i, \mathbf{a_t = e_{\ell}}\rbr  \nonumber \\
  & \qquad \qquad \qquad  \qquad  + \Pr \lbr \mathbf{a_t = 0 }\rbr \Pr \lbr{Y}_t | \mathbf{a_t = 0}\rbr  \nonumber\\
 &= \sum_{\ell = 1}^M \lambda_{\ell} \Pr \lbr y^{\ell}_t | X_t = i,  \mathbf{a_t = e_{\ell}} \rbr \prod_{m \neq \ell}^{M} \Pr \lbr y^{m}_t | \mathbf{a_t = e_{\ell}} \rbr \nonumber\\
 & \qquad \qquad \qquad \qquad +\lambda_{0} \Pr \lbr{Y}_t | \mathbf{a_t = 0}\rbr 
\label{eq:obs_like2}
\end{align}
In the first term of the final expression, the outer sum checks all possible assignments of detections to a target in state $i$, which are disjoint events. The inner product is simply a factorisation of  $\Pr \lbr Y_t |X_t, \mathbf{a_t} \rbr$  due to the conditionally independent observations, given the association $\mathbf{a_t}$.

Analogously to the single observation model, for each $y^m_t$, we define a likelihood function given $\mathbf{a_t}$ as, $c^m_i(t) = \Pr \lbr y^m_t | X_t = i, \mathbf{a_t} \rbr$, $\forall \ i,m,t$. If $ \mathbf{a_t = e_m}$ then this likelihood is of the same form as in the previous section, since the association vector is pointing to observation $y^m_t$. The two other cases, when $ \mathbf{a_t \neq e_m}$ and $ \mathbf{a_t = 0}$, we define as follows, assuming once again that non-target detections are generated in a similar way to target detections and each can be modelled via up to $M$ independent clutter processes $U_t$
\begin{align*}
\Pr \lbr y^m_t | \mathbf{a_t = e_{\ell} }\rbr &= \sum_{j=1}^N \Pr \lbr y^m_t | U_t = j,\mathbf{a_t = e_{\ell}} \rbr \Pr \lbr U_t = j \rbr \nonumber \\
& = \sum_{j=1}^N c^m_j(t) \Pr \lbr U_t = j \rbr  \\
\Pr \lbr Y_t | \mathbf{a_t = 0}\rbr &= \prod_{m = 1}^{M} \Pr \lbr y^{m}_t | \mathbf{a_t = 0} \rbr \nonumber \\
& = \prod_{m = 1}^{M} \sum_{j=1}^N \Pr \lbr y^{m}_t | U_t = j, \mathbf{a_t = 0} \rbr \Pr \lbr U_t = j \rbr
\end{align*}

Numerical examples of the multiple observation model above will be used in the numerical simulation section, with example realisations plotted. However, it is perhaps instructive to consider the noiseless case. Consider the case of 2 observations $Y_t = \lbr y^1_t, y^2_t \rbr$, this observation vector could correspond to 
\begin{eqnarray*}
\begin{bmatrix}
  y^1_t       \\[0.3em]
   y^2_t             \end{bmatrix}
\rightarrow \begin{bmatrix}

     X_t  =y^1_t       \\[0.3em]
      U_t = y^2_t             \end{bmatrix} or
\begin{bmatrix}

     U_t = y^1_t        \\[0.3em]
      X_t = y^2_t           \end{bmatrix},
\end{eqnarray*} 
 In this situation the underlying RC is still hidden, but only by ambiguity, since it is not known which observation in the list $Y_t$ corresponds to the target.

\section{Optimal Filtering for Hidden Reciprocal Chains\label{sec:filt}}
In this section, we derive optimal filters for the estimation of the state sequence of a HRC, and for evaluation of the likelihood of a HRC observation sequence. Based on the construction of the observation models, the filter takes the same form regardless of the observation scenario, and so using the HRC filter in either simply becomes a matter of calculating and using the different observation likelihood functions. Therefore we present the HRC filter for a general observation sequence model.

\input{filters_v8.tex}

 \subsection{Optimal Single and Multi-Observation HRC filters \label{sec:multiObs_filt}}
 
As stated, in the case where the observations are generated via the single or multi-observation models of \ref{ssec:single_obs}, we simply use the correct corresponding observation likelihood, ie. \eqref{eq:singl_obs_defn} or \eqref{eq:obs_like2}, and substitute these likelihoods into the Markov bridge APPs of Equation \eqref{eq:std_filt}.

 Note that the complexity of this filter is no greater in the single observation scenario, however in the multi-observation case the complexity of this filter is $O(N^3T + M^2N^2T)$, where the latter term is the burden of computing the likelihoods for the list of $M$ observations at each time $t$. Thus if $M<<N$, the complexity of the filter is dominated by the original term on the left, $O(N^3T)$. 


\section{Track Extraction Detectors\label{sec:det}}

The observation models we have presented, which represent the sensor level detection process, along with the associated filters, allow us to investigate the track extraction benefits gained by incorporating source-destination awareness into a detector's target model. We do this by defining several track extraction detectors, where each is a hypothesis test with two competing hypotheses. Based on the filters derived in the previous section we are able to formulate a likelihood ratio test in each scenario. Note that since an analytic form for the probability distribution of the sequence log likelihood under the null or alternative hypotheses can not be found, numerical approaches (such as simulations) are required in order to sensibly set the detection threshold.

The first class of detector we present attempts to decide whether a target exists in the cluttered environment. Thus in the single observation case the hypotheses are concerned with the value of $\epsilon$ in the mixture model. The null hypothesis is that the observations are all clutter generated ($\epsilon=1$). The alternative hypothesis presumes there is a reciprocal target, and over many realisations we expect to receive target generated observations at a rate ($1-\epsilon$).
\begin{eqnarray*}
H_0: & \epsilon = 1 & \text{(no target present)}\\
H_R: & \epsilon < 1 & \text{(a reciprocal target is present)}
\end{eqnarray*}
We call this a {\it reciprocal detector}. To compare the reciprocal model with a compatible Markov model, we form another detector where the alternative hypothesis instead has sequence log likelihood obtained from a standard HMC filter which assumes a target model with Markov dynamics.
\begin{eqnarray*}
H_0: & \epsilon = 1 & \text{(no target present)}\\
H_M: & \epsilon < 1 & \text{(a Markov target is present)}
\end{eqnarray*}
The dynamics could be those of the base Markov process, or of the Schr\"{o}dinger bridge with dynamics given by Equation \eqref{eq:SB_trans}. We call this detector a {\it Markov detector} if the base process is used to model dynamics, or a \emph{Schr\"{o}dinger detector} if a SB is used.

We also define corresponding detectors for the multi-observation case, which we parameterise via the priors over the association variable $\mathbf{a_t}$. Specifically, from \eqref{eq:multi_prior} the null hypothesis will be parameterised as $\lambda_0 = 1$, and the alternative hypothesis as $\lambda_0 <1$. The multi-observation likelihood for the null hypothesis is given by

\begin{align*}
\Pr \lbr {\Yc} | H_0 \rbr &= \prod_{t=0}^{T}\Pr \lbr Y_t | H_0 \rbr \\
& = \prod_{t=0}^{T}\prod_{\ell =1}^{M} \sum_{i = 1}^{N} \Pr \lbr y_{t}^{\ell} | U_t = i, \mathbf{a_t = 0} \rbr \Pr \lbr U_t = i \rbr.
\end{align*}

As before, we will define the temporally independent clutter process $U_t$ to be uniform over the state space. In the case of noiseless observations, the observation sequence likelihood expression above simplifies further.


\section{Numerical Examples \label{sec:sims}}

In this section, we present numerical simulations to illustrate the performance of the proposed filtering algorithms for a HRC track extraction scheme. The simulated scenarios are designed to highlight the benefit that can be achieved in both state estimation and detection when source-destination awareness is incorporated into the target model. We compare the HRC tracker (estimator and detector) to existing Hidden Markov chain (HMC) trackers similar to the track extraction models of \cite{vK,WK}, and also compare to a tracker based on a Schr\"{o}dinger bridge, which we call a Hidden Schr\"{o}dinger Chain (HSC) tracker.

The HMC tracker uses the base process defined in subsection \ref{ssec:encode_dest} as its target model, and the HRC and HSC trackers use target models with dynamics derived from this same base process. The models differ in terms of the statistical characterisation of the end-points and their dynamics ; the HRC has a specified joint distribution $\Pi$ on the end-points, the HMC has only initial state distribution $\pi_0$, being the marginal derived from $\Pi$ ; and the HSC has initial and final distributions equal to the marginal distributions of $\Pi$.

In all simulations the target trajectory data is generated according to the RC target model, for different distributions $\Pi$, and thus the HMC and HSC filters and detectors are mismatched while the HRC is matched. This is in line with the premise of the paper, that targets on courser time scales have at least a simple notion of intent, in this case source-destination awareness.

Firstly, in subsection \ref{ssec:simEnv} we briefly describe the simulation environment. In \ref{ssec:endptDist} we describe and justify the kinds of source-destination awareness we have chosen to encode with the RC model. In \ref{ssec:results}, we present the detection and filtering results, for both single and multiple observation scenarios, before we discuss these results in \ref{ssec:discussion}.

\subsection{Simulation Environment\label{ssec:simEnv}}

\begin{figure}[!h]
\begin{center}
\includegraphics[scale=0.55]{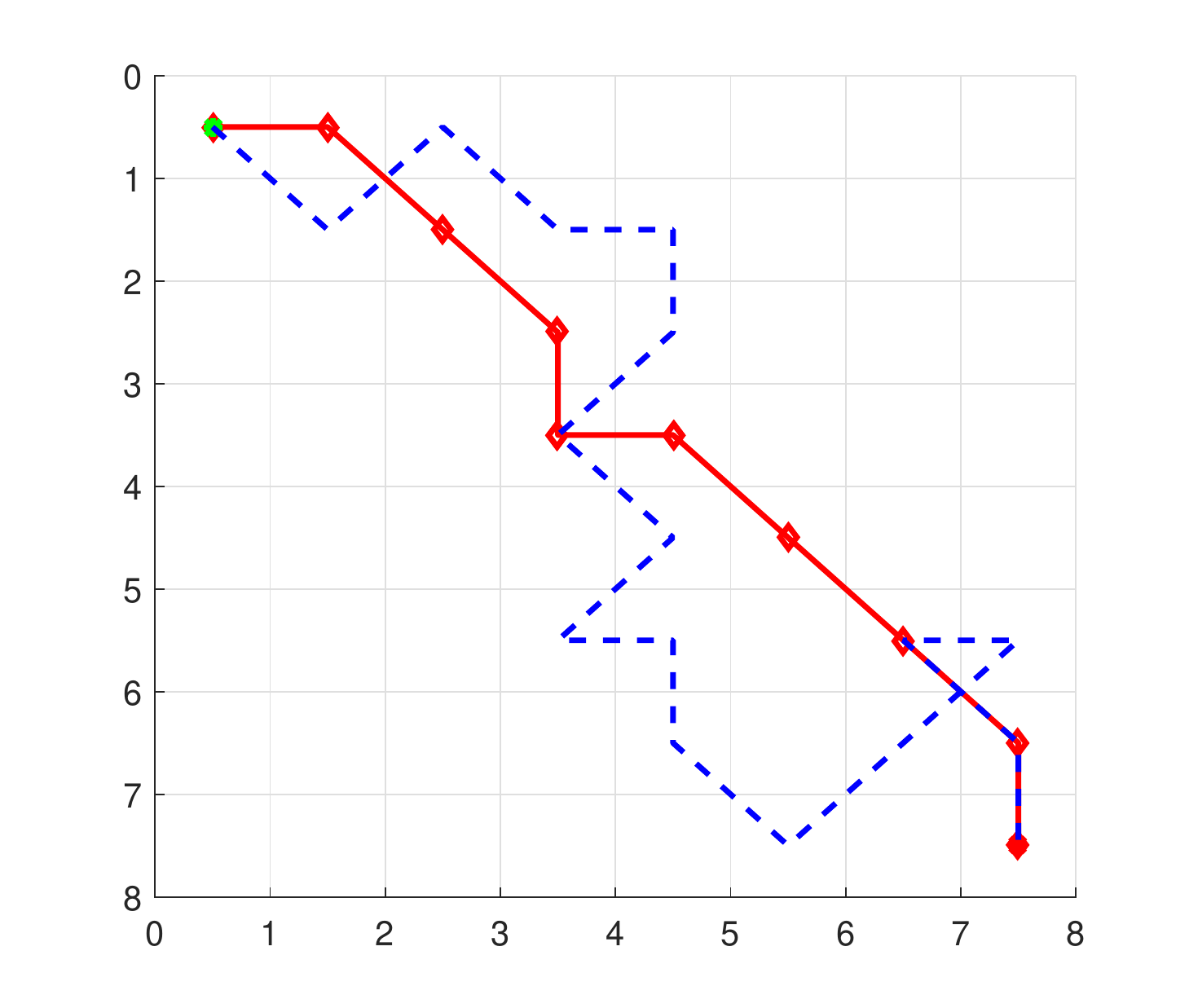}
\caption{ {\it Simple source destination aware trajectories}: The paths of 2 RC targets which cross the $8 \times 8$ lattice from $(1,1)$ to $(8,8)$. Green markers denote the origin state of any realisation, and the red denote the end state. Each trajectory corresponds to that of a Markov bridge with a fixed start, the simplest type of destination awareness. The red target has $T = 12$ steps to reach state $(8,8)$, while the blue has $T = 32$.\label{fig:graph_sim0}}
\end{center}
\end{figure}

In all the examples presented, the states take values on a regular two-dimensional $8\times 8$ cellular gridworld - the {\it region of interest} (ROI). We restrict our RCs to those whose local dynamics are built from one-step Markovian random walks, parameterised by the probability of remaining in a cell $p_R$ and equal probabilities of moving to the neighbouring cells (states). Note that neighbouring cells include those on the diagonal, meaning the random walk is `8-connected', rather than 4-connected, though this is an arbitrary choice. Jumps outside the ROI are not permitted. Fig. \ref{fig:graph_sim0} illustrates the basic nature of the state space and some sample trajectories.

The trajectories are realisations of two different RCs. Note that the plotted paths do not show if a target remains in or revisits a particular state. In order to generate an observation sequence, the independent clutter process $U_t$ is also realised. A sensor detection in the observation sequence is obtained by adding zero-mean Gaussian noise to the centre coordinates of the cell that the target or clutter is in at time $t$. Note this choice is illustrative and easily be generalised. In the single observation model, whether the target or clutter selected of course depends of course on the clutter rate $\epsilon$, and similarly in the multi-observation case. The added Gaussian noise has equal variance $\sigma^2$ in the $x$ and $y$ directions with the $x$ and $y$ components of the noise being statistically independent. Fig. \ref{fig:graph_sim3} presents an RC and the observation process in the single observation scenario.

\subsection{Joint Endpoint Distributions\label{ssec:endptDist}}

The performance of all of the detectors and filters considered depends on the ability of the underlying target model to accurately describe the target's dynamics. As stated, if there is a target, it is reciprocal. It would be expected that the HRC based detector should perform the best since it is matched to the data, but the performance benefit may be small, depending on how the destination awareness affects the RC dynamics, which are built from, and therefore different to, those of the base process. Therefore our numerical studies have focused on the parameters that determine the difference between the RC dynamics and the base process dynamics.

The red path, in Fig. \ref{fig:graph_sim0}, shows a target moving a distance of $7$ steps in an interval length of $T = 12$, whereas the blue target has $T=32$ to move the same distance. Both targets are realisations from RCs with the same $\Pi$ and dynamics built from the same MC dynamics $A$, but their trajectories are very different. The red target is more restricted and moves with a clearer purpose, while the blue target appears to move more randomly, with dynamics closer to the original Markov dynamics.

In general, it is not just the sequence length $T$ that determines how restricted or random the average dynamics of a RC target will be. This will also depend on  a) the size of the state space and the ease with which a target can traverse the lattice (whether it can only move one step or more, and with what likelihoods), which are both encoded into the transition matrix $A$, and in terms of average dynamics b) the distribution over the source-destination pairs, $\Pi$. Since for these simulations the state space and the transition matrix $A$ are fixed (and restricted to one step dynamics), we propose an empirical formula that relates the extent to which the RC dynamics are less random than the MC dynamics, to the free parameters: the interval length  $T$, the minimum number of steps $d_{min}(i,j)$ between source-destination pairs $(i,j)$, and the distribution over pairs $\Pi$. 

\begin{align}
\beta =   \frac{\sum_{i , j = 1}^N d_{min}(i,j) \Pi(i,j)}{T} \in [0,1] \label{eq:beta}
\end{align}

Based on intuition and general observation from simulations, we claim that the term $\beta$ will in turn predict the benefit that HRC will provide over Markov trackers according to a linear relationship, with a constant offset $\kappa$,

\begin{align}
\text{HRC Benefit} & \propto \beta +\kappa \label{eq:benefit_dest}
\end{align}

We will refer to $\beta$ as the benefit indicator, and define benefit in the following subsection, in the detection case. Notionally if $\beta$ is closer to $0$ we expect the RC dynamics to be similar to the MC dynamics, whereas if $\beta$ is unity, the RC dynamics result in a deterministic path and thus would appear quite different to the MC dynamics. We will present results testing this claim in the following subsection.

As an example, consider a RC model which only allows targets to start in the corners and cross the $8\times 8 $ state space to its opposite corner, as in figure \ref{fig:graph_sim1}. We will call endpoint distributions of this form a crossing RC (CRC). If we then set $T = 8$, the targets then have the minimum time required to move the $d_{min}(i,j) = 7$ steps between its start and final state. This motion is deterministic according to the RC, and while possible according to the MC, highly unlikely. The deviation for this case is $\beta = 1$. We would expect HRC trackers to perform best for target behaviour of this type. Another interesting test case is for when the target is still non-Markov but the benefit indicator $\beta$ is very low, or $0$. An example of this case is the loitering RC , for which the indicator is indeed $\beta = 0$, since the minimum number of steps $d_{min}(i,j) = 0$ (as $i=j$ for all source-destination pairs, see Eq. \eqref{eq:loitering_ends}).

We have chosen joint endpoint distributions $\Pi$ which range between the two cases given, crossing and loitering, according to the following construction
\begin{align}
\text{System } \Pi & =   \alpha \Pi_{CRC}+ (1- \alpha)  \Pi_{LRC} \ ,
 \label{eq:system_endpoints}
\end{align}
where $\alpha \in [0,1]$, and LRC denotes a loitering RC as before. We have used this construction to study the performance of the filters and detectors more systematically. Thus, as a consequence of the claim in equation \eqref{eq:beta}, it should be expected that the performance benefit can be controlled by the sequence length $T$ and the parameter $\alpha$, which ranges from what we expect to be best case ($\alpha = 1 $) to worst case ($\alpha = 0 $). We can express the benefit indicator as a function of $\alpha$,
\begin{align}
\beta =  \frac{\alpha}{T}\ d_{CRC}  \label{eq:beta_alpha}
\end{align}
since $d_{min} = 0$ for all loitering source-destination pairs, and assuming the same, constant number of steps $d_{CRC}$ for all crossing source-destination pairs. In the $8 \times 8$ state space $d_{CRC}= 7$ steps. Note that if $\alpha = 0 $, and all trajectories are loitering, $T$ can be lowered such that the end state is always reachable for any $T$.

\subsection{Results\label{ssec:results}}
In this subsection we present results for both the detection and filtering algorithms in the simulation environment described. All three trackers are applied in the single and multi-observation scenarios, with the results consistent across both scenarios. We begin first with results on detection, specifically with a focus on testing the claims of Equations \eqref{eq:beta} and \eqref{eq:benefit_dest}, when parameterised by $\alpha$ as in Equation \eqref{eq:system_endpoints}. We then present results on the estimation performance of the filters. Despite track extraction being largely about detection, the filtering results give us an insight into the detection mechanism, and why the different models perform differently.

Results on the effect of clutter and general measurement noise reflect the expected property that the benefit of better models increases with worsening tracking conditions. We provide evidence of this by comparing HMC to HRC performance as conditions worsen, strengthening the general argument made in the paper. We leave $A$ constant as mentioned for the primary detection and filtering simulations with the probability of the remaining in a cell fixed to $p_R = 0.5$. Following the primary results we will vary $p_R$ to investigate its effect on performance. Note that unless otherwise stated, results are obtained from $10,000$ independent realisations.

\subsubsection*{Detection}\quad

\begin{figure}[!h]
\begin{center}
\subfigure[]{\label{fig:graph_sim1a}
\includegraphics[scale=0.65]{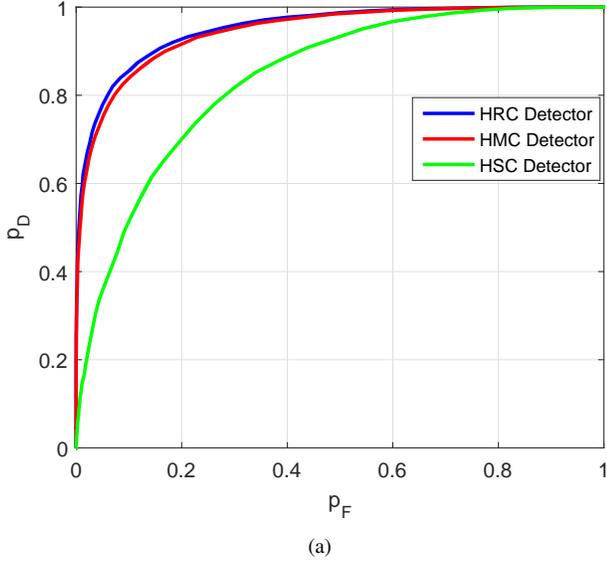}}
\subfigure[]{\label{fig:graph_sim1b}
\includegraphics[scale=0.65]{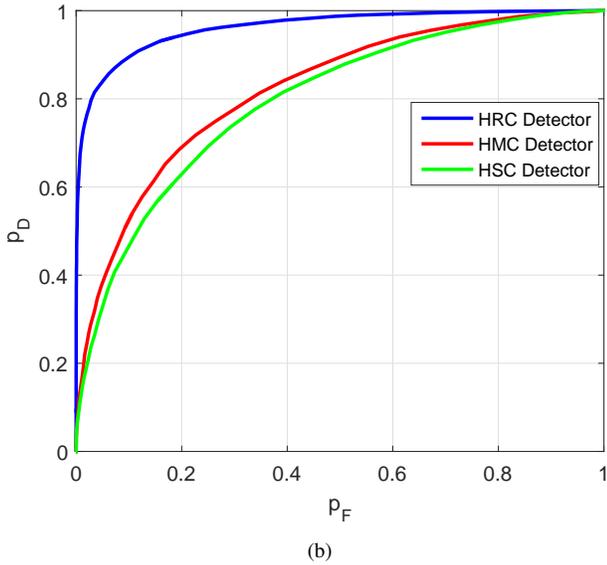}}
\caption{ {\it Single Observation Detection Results}: ROC curve for a single observation system with clutter rate $\epsilon = 50\%$ and sensor detection noise $\sigma^2 = 1$. The RC has interval length $T = 16$, and in a) $\alpha = 0$, b) $\alpha =1$.\label{fig:graph_sim1} }
\end{center}
\end{figure}

Detection results are presented using receiver operator characteristic (ROC) curves to begin with in Figure \ref{fig:graph_sim1}, which plot the estimated probability of detection against the estimated false alarm rate as the detection threshold $\tau$ is varied. We simulate realisations of both hypotheses in equal number, RC target present and no target present, reflecting the uniform priors chosen. The detectors perform the likelihood ratio test using the uniform priors, equal penalties for incorrect decisions and no penalties for correct decisions, that is, a minimum error probability test.

To begin with we consider the single observation scenario with a clutter rate $\epsilon = 50\%$ and increase $\alpha$ from $0$ to $1$. This means we should expect to see the benefit of the RC increase with $\alpha$, which is shown in Fig. \ref{fig:graph_sim1}. The multi-observation case obtains similar results across all $\alpha$ settings.

\begin{figure}[!h]
\begin{center}
\includegraphics[scale=0.6]{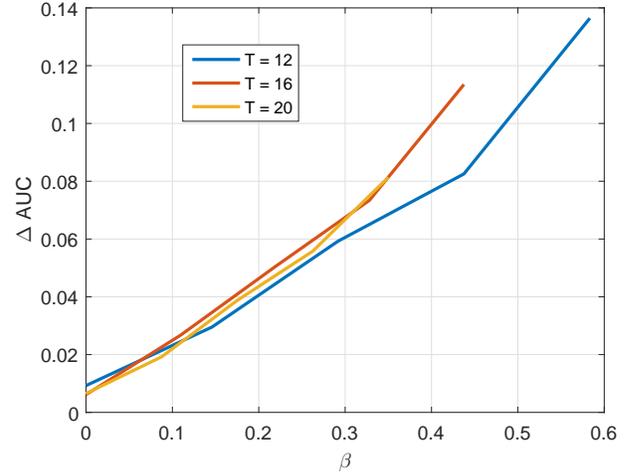}
\caption{ {\it $\Delta AUC$ vs. $\beta$}: Benefit provided by RC taken to be area between ROC curves oh HRC and HMC ($\Delta AUC$) in a single observation scenario, with $\epsilon = 0.4$, $\sigma^2 = 1$, $T = 16$ as $\alpha$ increases. \label{fig:graph_sim2}}
\end{center}
\end{figure}

In Figure \ref{fig:graph_sim2} we present results testing the validity of the proposed relationship and formula in \eqref{eq:benefit_dest}, with the area under the curve (AUC) of the ROC curve taken to be the measure of benefit that HRC provides over the HMC tracker. The AUC is a valid measure of aggregate detector performance provided that one of the detectors being compared has ROC curve consistently above the other curve \cite{H}, which is the case here. The benefit indicator $\beta$ is varied via \eqref{eq:beta_alpha} as $\alpha$ increases from $0$ to $1$, for different fixed interval lengths $T$, corresponding to each curve. We can see that the proportional relationship is approximately correct, especially as sequence length $T$ increases. We see also that the curves become shorter, which is to be expected according to Equations \eqref{eq:beta} and \eqref{eq:benefit_dest}.

\subsubsection*{Filtering}\quad

\begin{figure}[!h]
\begin{center}
\includegraphics[scale=0.6]{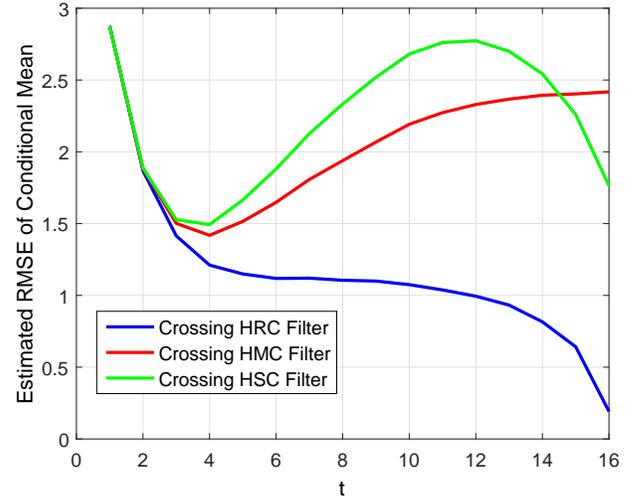}
\caption{ {\it Single Observation Filtering Results}: Estimated RMSE of the Conditional Mean for each of the trackers with the clutter rate of $\epsilon = 0.25$, and sensor detector noise of $\sigma^2 = 1$  . The RC has interval length $T = 16$,, and $\alpha = 1$. \label{fig:graph_sim3}}
\end{center}
\end{figure}

In the filtering example of Figure \ref{fig:graph_sim3}, we set $\alpha = 1$ and purely considers CRC targets, with a path length of $T=16$, in the single observation scenario. The clutter rate was set to $25\%$ ($\epsilon = 0.25$), meaning that on average $75\%$ of any observation sequence will be of target generated measurements.

The performance of the filters is measured by the RMSE of the online conditional means. The conditional mean is a continuous real valued estimate corresponding to the $2D$ gridworld state space,
\begin{align*}
\hat{X}_t = \Er \lbr X_t|Y_0,\dots Y_t\rbr
\end{align*}
Given $X_t^r, Y_t^r, \hat{X}^r_t$ are the states, observations and conditional mean estimates corresponding to a particular realisation $r \in \lbr1,\dots,R\rbr$ at each time $t$, the RMSE of the conditional mean estimates is calculated using the Euclidean distance from the target's state coordinates, $x^r_t, y^r_t$ to the estimate coordinates, $\hat{x}^r_t, \hat{y}^r_t$,
\begin{align*}
\text{RMSE}_{CM}(t) = \sqrt{ \sum_{r = 1}^R \frac{(\hat{ x}_t^r - x^r_t)^2 + (\hat{ y}_t^r -  y^r_t)^2}{R} }
\end{align*}
where. The RMSE of the conditional mean provides a good measure of tracking accuracy. The results for the multi-observation scenario are very similar for online estimation, to within a small scaling factor. 

\begin{figure}[!h]
\begin{center}
\includegraphics[scale=0.6]{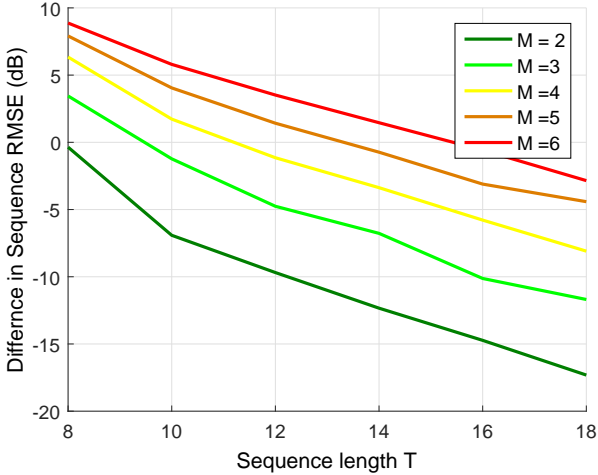}
\caption{ {\it Multi Observation Filtering Results}:  Filtering performance difference of HRC over HMC model. The vertical axis is the benefit in RMSE in dB of the average per sample RMSE as interval length $T$ is varied, for different number of observations  per time $M$, from $M = 2$ (dark green) to $M = 6$ (red). The horizontal axis is the sequence length $T$. \label{fig:filt_sim4}}
\end{center}
\end{figure}

As stated, we expect that HRC performance benefit increases with worsening tracking conditions, since the model information becomes more useful. Figure \ref{fig:filt_sim4} shows the difference in average per sample RMSE between the HMC and HRC filters, in the multi-observation scenario between, as the number of observations per time increases. We define the average per sample (APS) RMSE of the filters as,
\begin{align*}
\text{RMSE}_{APS} = \sum_{r = 1}^R \frac{1}{R} \sqrt{\sum_{t=1}^T \frac{(\hat{ x}_t^r - x^r_t)^2  +  (\hat{ y}_t^r -  y^r_t)^2}{ T}}
\end{align*}
This produces a similar measure to the RMSE defined previously, but across the entire sequence, so that we may describe performance as the interval length $T$ is varied.
Each curve in Figure \ref{fig:filt_sim4} represents a different set of independent realisations corresponding to the number of observations at each time, increasing from $M = 2$ to $6$, when all targets are crossing. We set the \emph{a priori} probability of no target detection, to $\lambda_0 = 0$ (see Equation \eqref{eq:multi_prior}), since the effect of $\lambda_0 >0$ is seen in the single observation scenario, thus the multi-observation filter is resolving ambiguity. The phenomena of increasing performance of HRC over alternative Markov models was observed across all worsening conditions; increased sensor detector noise $\sigma^2$,  clutter rate $\epsilon$ as well as the number of targets $M$. We note also that the benefit of HRC over HMC decreases with increasing sequence length $T$.

\subsubsection*{Varying transition probabilities of $A$}\quad

\begin{figure}[!h]
\begin{center}
\subfigure[]{\label{graph_p_crossA}
\includegraphics[scale=0.6]{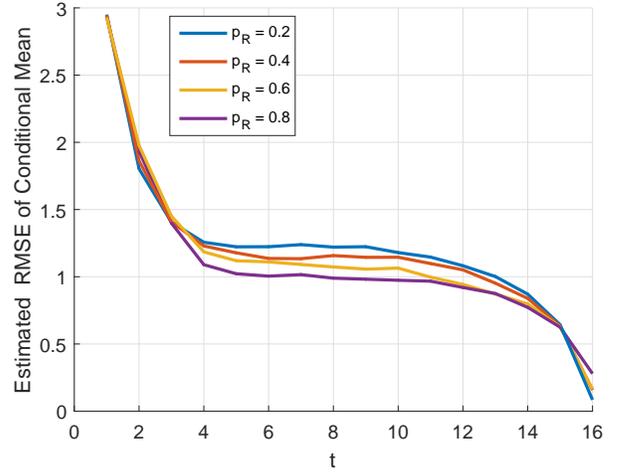} \label{fig:piHRCcross}}
\subfigure[]{\label{graph_p_crossB}
\includegraphics[scale=0.6]{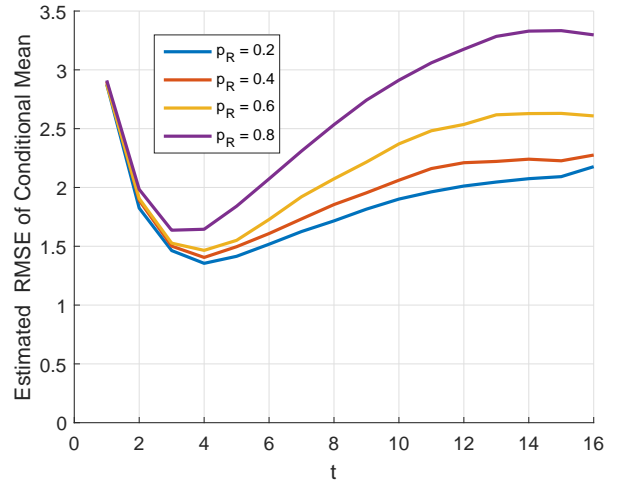} \label{fig:piHMCcross}}
\caption{ {\it Filtering conditional mean RMSE for different $p_R$}: target is a crossing RC ($\alpha = 1$), $\epsilon = 0.5$, $T = 16$, a) HRC b) HMC. \label{fig:piCross}}
\end{center}
\end{figure}

\begin{figure}[!h]
\begin{center}
\subfigure[]{\label{graph_p_crossA}
\includegraphics[scale=0.6]{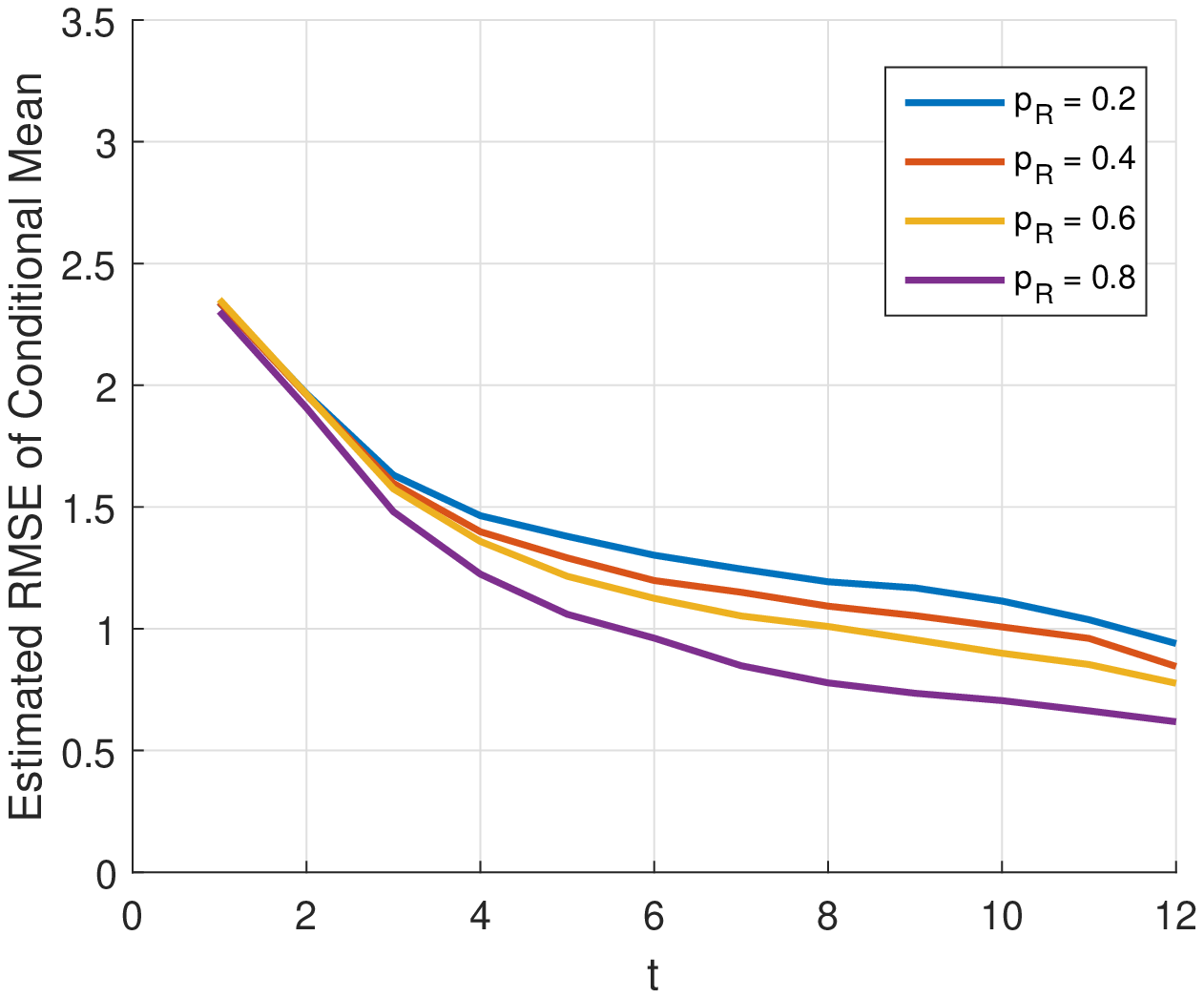} \label{fig:piHRCloit}}
\subfigure[]{\label{graph_p_crossB}
\includegraphics[scale=0.6]{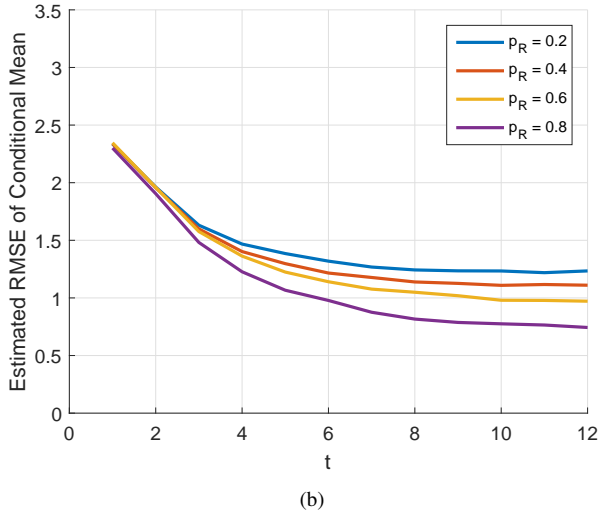} \label{fig:piHMCloit}}
\caption{ {\it Filtering conditional mean RMSE for different $p_R$}: target is a loitering RC ($\alpha = 0$), $\epsilon = 0.5$, $T = 12$, a) HRC b) HMC. \label{fig:piLoit}}
\end{center}
\end{figure}

The final numerical simulations investigate the effect that varying the one-step transition probabilities $A$ of the base process have on the HRC and HMC trackers' performance, using RMSE of the conditional mean estimates to measure the performance once more. We omit the HSC tracker for reasons that will become clear in the discussion. Figure \ref{fig:piCross} shows the performance of the HRC and HMC filters respectively, in the single observation scenario, with only CRC targets ($\alpha = 1$) as the probability of remaining in a vertex $p_R$ is varied from $0.2$ to $0.8$. We see that the HRC is largely invariant to the variation in $p_R$, while the HMC is not, estimating more poorly for higher $p_R$. Figure \ref{fig:piLoit} however, shows no difference between the HRC and HMC filtering performance when all the targets loiter.

\begin{figure}[!h]
\begin{center}
\includegraphics[scale=0.55]{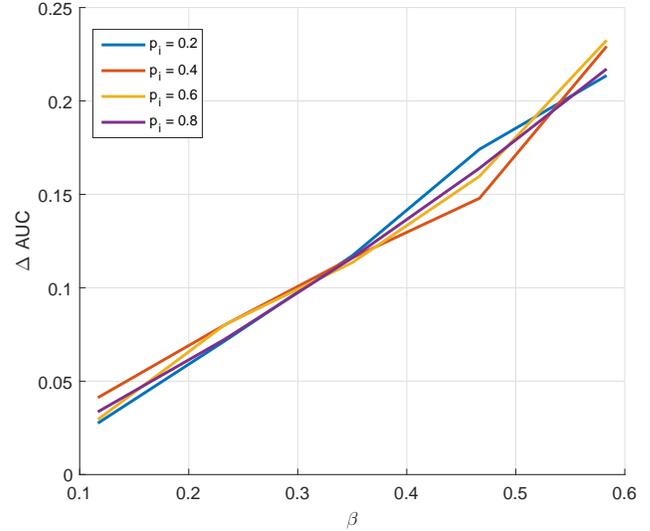}
\caption{ {\it $\Delta AUC$ vs. $\beta$ for different $p_R$}: Benefit once more taken to be the difference in area under the ROC curves of the HRC and HMC trackers ($\Delta AUC$). $T = 16$ as $\alpha$ is varied, $\epsilon = 0.5$. Graph shows invariance with respect to $p_R$. \label{fig:auc_diff_pi}}
\end{center}
\end{figure}

This effect of $p_R$ on the HRC and HMC trackers does not carry through from filtering to detection. We see this in Figure \ref{fig:auc_diff_pi}, where the difference in AUC between the HMC and HRC ROC curves is not effected by the change in $p_R$.

\subsection{Discussion\label{ssec:discussion}}

The results show that the HRC tracker, which is matched to the model generating realisations, performs best when compared to the Markov trackers, as expected. Furthermore the proposed formula \eqref{eq:beta} is supported, since the claim of linear proportionality \eqref{eq:benefit_dest} appears to be approximately correct, especially for longer $T$.  Despite lacking a direct probabilistic interpretation, Equation \eqref{eq:beta} provides a good intuitive description of the relationship between the parameters of the HRC model and the benefit over the HMC model.

When we consider the HMC and HSC results specifically we discover that although the Schr\"{o}dinger bridge model incorporates more future information than the Markov model, it performs worse. The first reason for this is that the Schr\"{o}dinger bridge construction introduces source destination pairs not present in the original $\Pi$, as indicated in section \ref{ssec:encode_dest}. This can be understood by considering the endpoint marginals obtained from the RC and passed to the SB. In the loitering case for example if the RC target can loiter anywhere uniformly in the state space, this will result in uniform marginals across both start and end points for the SB. Thus the HSC tracker will generally `expect' trajectories that will never be realised from the RC model. The second reason for these errors is that the clutter observation model allows the HSC tracker to assign the majority of measurements to clutter. This effect is evident in both the detection and filtering results, such as Figures \ref{fig:graph_sim1b} and \ref{fig:graph_sim3}, and points to the advantage of RC over SB modelling of targets with future information.

\begin{figure}[!h]
\begin{center}
\includegraphics[scale=0.6]{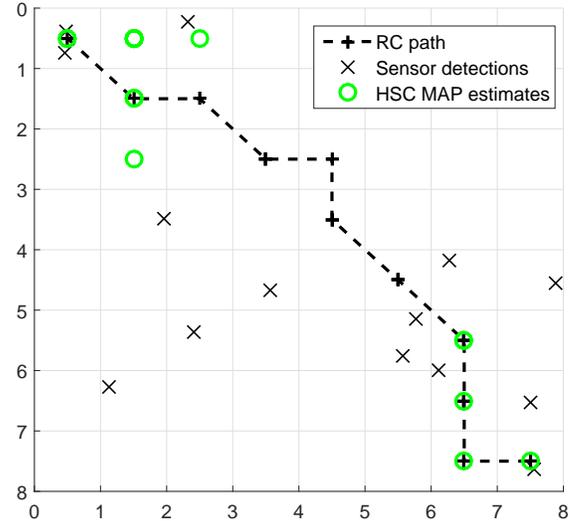}
\caption{ {\it HSC estimates predicting incorrect trajectory}: The figure above shows the target's path from $(1,1)$ to $(8,8)$ (dotted black line), the sensor detections (black points) and the HSC filter MAP estimates (green circles), for a crossing RC ($\alpha = 1$) with $T = 16$. Note there are in fact 16 points for each of the plots, however some overlap. It can be seen the HSC predicts a path close to the origin (state $1$, $(1,1)$) until $t=9$ when the last 4 data points are much closer to the target path.\label{fig:graph_sim8}}
\end{center}
\end{figure}

We can explain the HSC results further with a specific example of a crossing target in the single observation scenario. Figure \ref{fig:graph_sim8} is a plot of one particularly poor performance of the HSC tracker. The figure displays the target path $\Xc$, a noiseless path of the sensor detections (both target and clutter), the observations $\Yc$ and the HSC filter's maximum \emph{a posteriori} probability (MAP) estimates. Note there are $12$ HSC filter estimates (green), however many overlap. The majority of the HSC estimates are focused around the starting vertex $(1,1)$, since under its model it is possible for a target to start and end at the same location (due to the endpoint marginals). However, at time $t = 9$ the filter `realises' the measurement sequence makes more sense with a trajectory of $(1,1)$ to $(8,8)$. This effect corresponds to Figure \ref{fig:graph_sim2}, where the RMSE in the HSC conditional mean estimates decreases close to $T$.

Finally, we consider Figures \ref{fig:piCross} and \ref{fig:piLoit}, which present RMSE of the conditional mean of the filters as $p_R$ varies. Since a MC will have a distribution over its state at time $t$ (i.e. $\pi_0 (A)^t$ ) that is more concentrated about its origin for higher $p_R$, it makes sense that for higher $p_R$ the HMC should produce worse estimates for a CRC but better estimates for a LRC. By their definitions a CRC forces the target to move the furthest distance in time $T$ whereas a loitering RC alters the base process dynamics only to return to the origin with probability $1$, which explains the HMC performance. As state before however, this effect does not carry over to the detector performance.
 
Thus the two main results from the numerical studies are, primarily that the benefit of the HRC is controlled by the parameters of the RC model, and secondly that the HSC is a poor model for tracking targets that proceed from an initial state to arbitrary final state, and an analyst would be better served with a simple HMC tracker that uses the globally naive base process.

\section{Conclusion \label{sec:conc}}

This paper has considered the problem of how target models which incorporate a simple notion of intent can improve the tracking performance over alternative Markov target models. To do this, track extraction scenarios were recreated by defining two observation models of detections generated by thresholding sensor returns, with targets being modelled as reciprocal chains and clutter as a temporally uncorrelated process. A reciprocal chain can be constructed from a MC base process together with a joint endpoints distribution which encodes the source-destination awareness of a target. Normalised HRC filters were presented for the observation models introduced and the normalising constants were used to form likelihood based detectors. These HRC filters and detectors were compared via numerical simulations to both a HMC tracker with the base process as its target model, and a Markov tracker based on the Schr\"{o}dinger bridge, which can encode some future information.

Numerical simulation design was of a HRC on a 2D cellular gridworld of fixed size, with states as cells, and dynamics restricted to one-step walks over cells. The central result from these numerical simulations was that the benefit of the source-destination awareness of an RC, built from a base process with known one-step Markov dynamics, can be related to the sequence length, the distance between source and destination and the likelihood of that a target with that source and destination being realised. Specifically, if benefit is taken to be the difference in area under the ROC curves of the reciprocal and base Markov detector, it was found to vary approximately linearly with a simple function of the RC parameters in a specific scenario.

An important secondary result is that the Schr\"{o}dinger bridge tracker, despite having the correct start and endpoint distributions pre-specified, in fact performs worse than even a Markov tracker using base process.

Research tasks that follow directly from this work are first and foremost a generalisation of the central result of the simulations, with fewer restrictions on the base process (ie. allowing more than simple one-step dynamics) and subsequently replacing the empirical benefit indicator $\beta$ with a more general metric to indicate how the reciprocal dynamics deviate from Markov dynamics as a function of the RC parameters.
Most pressing research would be to address the problem of parameter estimation, and the related questions of how suitable the algorithms presented are in a real tracking system, and whether HRCs are strictly batch methods or could be feasibly employed online. Of the many parameters to be estimated, the most interesting and challenging is the joint endpoints distribution over the source and destinations, and how to select an appropriate corresponding $T$. In a real tracking environment there would be many targets behaving in various ways, with the potential for different RC targets existing at once. This research would benefit greatly from considering situations with corresponding real data, such as in \cite{PLFK}-\cite{ZMBD} and the references therein.

\section*{Acknowledgements}

The support of the Defence Science and Technology Group, Australia, and the Data to Decisions CRC is gratefully acknowledged. We would also like to thank Dr. Jason Williams for his useful insight and contribution.\\



\begin{IEEEbiography}[{\includegraphics[width=1.2in,height=1.25in,clip,keepaspectratio]{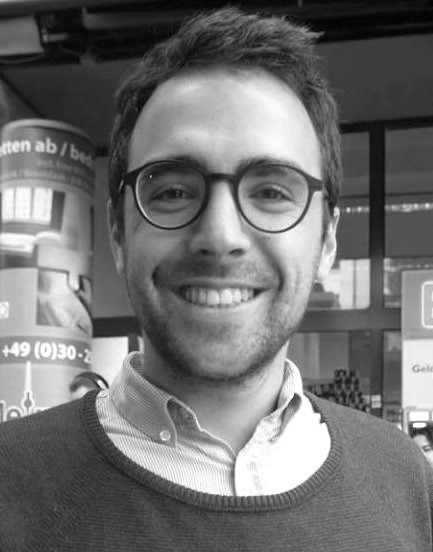}}]{George Stamatescu}
received the degrees of B. E. (hons) and B. Maths \& Comp. Sc. in 2013 from the University of Adelaide, where he is currently a graduate student with the School of Electrical and Electronic Engineering. His research interests lie in the application of finite state stochastic processes in tracking and artificial intelligence problems.
 \end{IEEEbiography} 
\begin{IEEEbiography}[{\includegraphics[width=1.1in,height=1.25in,clip,keepaspectratio]{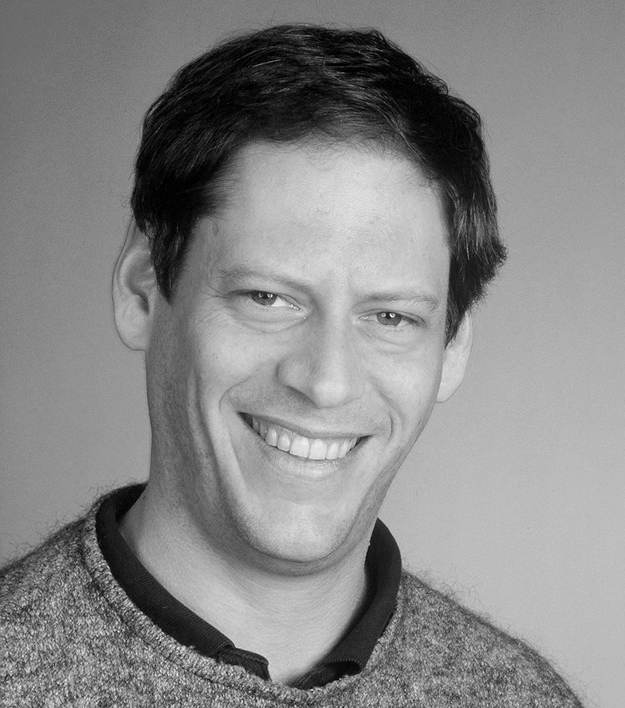}}]{Langford B white} obtained the degrees of B. Sc.
(mathematics), B. E. (hons) and Ph. D. degrees in
Electrical Engineering in 1984, ‘85 and ‘89 from
the University of Queensland. From 1986-1999 he
worked for the Defence Science and Technology
Organisation in South Australia. Since 1999 he has
been a Professor in the School of Electrical and
Electronic Engineering at the University of Adelaide.
Prof White’s research interests include signal processing,
control and game theory.
\end{IEEEbiography} 
\begin{IEEEbiography}[{\includegraphics[width=1.2in,height=1.25in,clip,keepaspectratio]{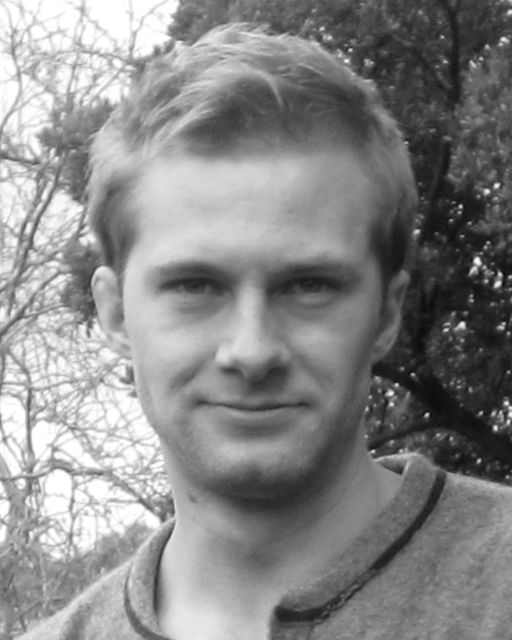}}]{Riley Bruce-Doust}
received the degrees of B. E. (hons) and B. Sc. in 2009 from the Australian National University. He is currently a graduate student with the School of Electrical and Electronic Engineering at the University of Adelaide. He is researching properties of finite-state stochastic processes. 
 \end{IEEEbiography}
\end{document}

%% file: intro_v8.tex
\subsubsection{Problem Description}
Track extraction is a statistical hypothesis testing technique that tests for the presence of a dynamic target in a set of detections of uncertain origin that are obtained from a detector applying a low threshold to sensor measurements \cite{vK, WK}. While standard multiple hypothesis tracking (MHT) algorithms confirm or delete tracks based on the likelihood of individual candidate tracks \cite{Blackman}, track extraction algorithms use the entire set of detections to first decide whether a target exists, with the likelihood supplied by applying a tracker with an underlying model of target dynamics. It is therefore also a detection process, with quantities such as detection probability and false alarm rate, but works on a higher level of abstraction than a sensor detector, leading to tracks rather than potentially unrelated point estimates. Classically, tracking and track extraction algorithms apply the Markov assumption to the target dynamics. At finer time scales the assumption is generally appropriate, and well-known Markov based tracking algorithms are discussed in the literature \cite{Blackman},\cite{BP}. This paper is motivated by tracking applications with coarser time scales such as \cite{PLFK}-\cite{ZMBD} where the Markov assumption is less valid generally because targets have a defined origin and destination. We will call such targets {\it source-destination aware}, and as in \cite{PLFK}-\cite{ZMBD} we work with finite-state processes because the arbitrary source-destination pair relationships are better captured by a cellular state space. In contrast to \cite{MDKL}-\cite{ZMBD}, which  largely focus on applications and only model a target's destination implicitly or via an ad hoc approach, we examine explicitly whether the use of dynamic target models which incorporate such information can improve track extraction.\\


\subsubsection{Related Work}
From a modelling point of view, being source-destination aware means that the initial and final target states have a specified joint probability distribution and that target dynamics are anticipative, or non-causal, reflecting the intention of the target to move towards its destination. There have been a range of approaches to incorporating available \emph{a priori} information about the future within the class of Markov models, both at the estimation stage and in the target dynamics model itself. Both methods amount to ``back propagating'' the influence of a future state value or distribution by appropriate conditioning. For example in \cite{CLW85},\cite{KP}, \emph{a priori} destination information is incorporated via a corrective term into the state update of continuous-time Gauss-Markov processes, and in \cite{Jami75}-\cite{GP} specifying an \emph{a priori distribution} for a future time of the state process creates a new Markovian process referred to as a Schr\"{o}dinger bridge. What is however not able to be modelled within the Markovian class is the future information that is conditional on the past state of the system (the intention to proceed between a source and associated destination). The injection of a relationship between future and past states other than that induced by the evolution of the “globally naive” first order Markovian dynamics, adds a probabilistic dependence that raises the class of model to what is known as the class of {\it reciprocal processes}. Reciprocal processes were studied in detail by Jamison \cite{Jami74} in a general setting, and subsequently by Levy {\it et al} \cite{LFK90} who considered the realisation and state estimation problem for the Gaussian discrete parameter case. There are many other related works which are summarised in \cite{WC11}, to which to the interested reader is referred for more background. \\

Reciprocal chains (RCs) are finite-state reciprocal processes, and hidden reciprocal chains (HRCs) are stochastic processes generated from an RC via some noisy and/or incomplete observation mechanism, analogous to hidden Markov chains (HMC). As stated we work with finite-state processes primarily because we are concerned with targets with arbitrary source-destination pair relationships and the precision of point estimates produced by continuous Gauss-Markov processes are not relevant. Finite states also allow for state space constraints, such as ``forbidden areas'' for the target or restrictions to roads as in \cite{PLFK}-\cite{ZMBD}, and more abstract target attribute states. We thus recount the present state of development in inference of HRCs; un-normalised optimal filters/smoothers for HRC, derived using a Bayesian approach, were presented in \cite{WC11}. Normalised filters and smoothers were developed in \cite{WC14}, which also considered a generalisation of HRC to incorporate ``waypoints'' on the target trajectory. Maximum likelihood estimation of state sequences (MLSE) for HRC was presented in \cite{WV13}. In \cite{FK13},\cite{FKW11} an approach called destination aware tracking based on HRC was first proposed, but the RC models tested, being pinned to a single destination, in fact remained in the Markov class. Thus the tracking performance of non-causal HRCs has not been studied in detail. The present paper is an extension of the work presented in \cite{SDW}, which provided an intuitive description of how HRCs may be applied in a target tracking problem. The filtering algorithms derived in this paper proceed in the manner of the Bayesian decomposition of a reciprocal chain into a collection of end-pinned bridge processes that was established in \cite{WC11}.

\subsubsection{Contributions}
The key contributions of this paper are in answering the questions of whether modelling the source-destination awareness of a target can improve track extraction, and to what extent it is improved as a function of the parameters of a HRC, which model source-destination awareness. 
The track extraction setting to which we introduce the new model, and compare to existing models, is similar to the setup in \cite{vK}, with three key differences. Firstly, we perform optimal Bayesian estimation over finite states, rather than a continuous space. Secondly we operate on a fixed interval rather than in an online context, and thirdly the likelihood ratio test here has a single threshold, though it could be formulated as the sequential probability ratio test of \cite{vK} which has two thresholds. To recreate the setting, we include two types of uncertainty, false alarms - detections that do not originate from the target (ie. from `clutter’), and an arbitrary `sensor' noise which degrades position estimates. This could noise could include the effect of quantisation of sensor measurements associated with the finite state space. In section V of \cite{vK}, signal-strength information is used and improves the track extraction performance, but here we remove any explicit notion of SNR and signal amplitude to emphasise the role of the model in differentiating clutter from target detections. Instead we treat the association between observation and the source of the detection probabilistically, via an observation likelihood function and \emph{a priori} probabilities over false alarms. Observation models for two representative track extraction regimes are developed and tested. In each there is at most one target during the tracking interval. In the first we process one sensor detection at each time increment, with a given fixed \emph{a priori} probability it is a false alarm. In the second scenario we generalise this to allow multiple detections at each time, with at most one due to the target, a scenario also typical in visual tracking and which is very similar to the observation model of \cite{vK}. A dynamic model of the target is constructed which incorporates target dynamics, including the global source-destination attributes. The likelihood associated with a given set of detections is evaluated using the new class of normalised hidden reciprocal chain (HRC) filters. This likelihood forms the basis for a target extraction algorithm.
With most finite data non-linear inference problems, it is infeasible to obtain {\it a priori} performance metrics, so we are led to the use of numerical simulations to study the performance of the new algorithms proposed here.

There has been a steady shift across tracking application domains towards modelling more complex target behaviours, for example in user support applications such as location based services (eg. \cite{PLFK},\cite{MDKL}), and in tracking scenarios where analysts' decision making can benefit from models that suit questions of higher level inference, such as predicting destination (eg. \cite{FK13}). The contributions of this paper are relevant to these application areas as they describe, via numerical studies, the tracking benefit that stands to be gained by incorporating a simple notion of intent. Alternative approaches \cite{PLFK}-\cite{ZMBD} have largely assumed rather than demonstrated that modelling intent improves tracking, which this paper focuses on explicitly within the track extraction setting.

\subsubsection{Paper Outline}

 The layout of the paper is as follows: in Section II, we define HRC and summarise the pinned Markov process construction of an HRC such as given in [6]. The observation models are defined in Section III before the normalised optimal filters are derived in Section IV, and the corresponding detectors in section V. Numerical simulations are presented in Section VI, before we conclude in Section VII with a discussion of future research directions.

%% file: MB_construction_v8.tex
Consider a random process $\{X_t\} $ 
indexed by $t \in \{0, 1,\ldots, T\} $ for some fixed integer $T \geq 2$. 
The index $t$ will denote an index on a set of event epochs. 
At each epoch $t$,  the random variable $X_t$ takes a value on a finite state space $\Sc = \{1,\ldots, N\}$, where $N \geq 2$ is a fixed integer.
The process $\{X_t\}$
is said to be {\it reciprocal} \cite{Jami74}, if 
\begin{eqnarray}
\Pr \lbr X_t | X_s, \forall \, s \neq t \rbr & = & \Pr \lbr X_t | X_{t-1},
X_{t+1} \rbr \ ,
\label{eq:rp1}
\end{eqnarray}
for each $t = 1, \ldots, T-1$. Thus $X_t$ is conditionally independent
of $X_0, \ldots, X_{t-2}, X_{t+2}, \ldots, X_T$ given its neighbours
$X_{t-1}$ and $X_{t+1}$. The reciprocal model is specified by the set of three-point
transition functions \eqref{eq:rp1} together with a given joint
distribution on the end points $\Pr \lbr X_1, X_T \rbr$. 
A reciprocal process is not, in general, a Markov process, however all Markov processes are reciprocal \cite{Jami74}.
Denote the three-point transition functions in (\ref{eq:rp1}) by
\begin{eqnarray}
Q_{i,j,\ell}(t) = \Pr \lbr X_t = j | X_{t-1}= i, X_{t+1} =
\ell \rbr),  \label{eq:3pt_tr}
\end{eqnarray}
for $i,j,\ell \in \Sc$, $t=1, \ldots, T-1$, and the end-points distribution given by
\begin{eqnarray}
\Pr \lbr X_0=i, X_T=j \rbr & = & \Pi_{i,j}, \quad i,j \in \Sc \ .
\label{eq:end_pts}
\end{eqnarray}
Pinning the end point of a RC generates a Markov bridge, which we consider to be a Markov process with the end point $X_T$ fixed to a specified value, together with the initial state distribution, obtained from \eqref{eq:end_pts}, conditioning on the value $X_T$ takes. The term bridge generally refers to a process with specified values at both start and end times, however provided the end is reachable one may define it with an initial distrubtion rather than fixed initial state.

So for the finite ($N$) state case, a RC can be
regarded as $N$ Markov bridges, one corresponding to each of the
possible final states taken by $X_T$. When we consider the joint distribution of the states of a RC and using direct Bayes' conditioning and \eqref{eq:rp1}, the relevance of Markov bridges becomes apparent,
\begin{align}
\Pr \lbr X_0, \ldots, X_T \rbr & = \Pr \lbr X_1, X_T \rbr \lb \prod_{t=2}^{T-1} \Pr \lbr X_t |
X_{t-1},X_T \rbr \rb .
\label{eq:RC_like}
\end{align}
The terms contained within the product in \eqref{eq:RC_like} are precisely the state transitions for a MB pinned at $X_T$, and depend only on the three-point transitions \eqref{eq:3pt_tr}. Based on the properties of a RC a
backwards recursion was obtained in \cite{WC11} which fully specifies the set of $N$ MB transitions for $t=T-2,T-3, \ldots,
1$ via
\begin{align}
B_{i,j}^k(t)  & = \Pr \lbr X_{t+1} = j | X_t = i, X_T = k \rbr \nonumber \\ 
&=  \frac{Q_{i,j,\ell}(t+1)}{B_{j,\ell}^k(t+1)} \lb 
\sum_{m=1}^{N} \frac{Q_{i,m,\ell}(t+1)}{B_{m,\ell}^k(t+1)} \rb^{-1}
,\  \label{eq:MBform2}
\end{align}
the last term on the right being the normalisation constant. Initialisation
is with $B_{i,j}^k(T-2) = Q_{i,j,k}(T-1)$. The quantity
on the right hand side of equation \eqref{eq:MBform2} is independent of the index
$\ell$. Now since the terms
$B_{j,.}^k(t+1)$ form a probability distribution (i.e. they are
non-negative and sum to unity) for each $i,k$ and $t$, there is at
least one index $\ell$ for which $B_{j,\ell}^k(t+1)$ is non-zero. So
this index may be selected on the rhs of \eqref{eq:MBform2}, and thus
\eqref{eq:MBform2} is well defined. From a numerical perspective, it
may thus be appropriate to choose a value of $\ell$ (for each $j,k, t$) which
maximises the value of $B_{j,\ell}^k(t+1)$ when performing the calculations in \eqref{eq:MBform2}.  Determining the set of MBs from
the specified three point transitions requires complexity of
$O(N^3T)$, i.e. $O(N^2T)$ for each MB. The Markov bridge with final state $k$ has an initial
probability distribution $\pi^k_i$ given by the conditional
distribution
\begin{align}
\pi^k_i & = \Pr \lbr X_0 = i | X_T = k \rbr \ = \ \frac{\Pi_{i,k}}{\sum_{j=1}^N \Pi_{j,k}} \ .
\label{eq:MBinit}
\end{align}
Thus we have that any RC may be uniquely specified by the
finite set of Markov bridges with probability transition matrices
given by \eqref{eq:MBform2} and initial distributions
\eqref{eq:MBinit}. 
As we will see below, a more intuitive description of RCs can be found via the construction of a reciprocal process from a Markov process, which we will refer to as the {\it base process}.

%% file: HRCs_ssec_v7.tex
Suppose that the RC $\Xc = \{X_0,,\ldots,X_T\}$ is observed via the
 observation process $\Yc = \lbr Y_0, \ldots,
Y_T\rbr$.  Assume that the observation at epoch $t$ given the state $X_t$ is conditionally dependent of $X_\tau$ and $Y_\tau$,  $\tau \neq t$. 
This conditional independence implies that
\begin{eqnarray}
\Pr \lbr Y_0, \ldots, Y_T | X_0, \ldots, X_T \rbr & = &
\prod_{t=0}^T \Pr \lbr Y_t | X_t \rbr .
\label{eq:cond_ind}
\end{eqnarray}
The process $\Yc$ is called a hidden reciprocal chain (HRC) because the property
\eqref{eq:cond_ind} is analogous to the usual assumption made for
hidden Markov chains. 
Let
\begin{eqnarray}
C_i(t) & = & \Pr \lbr Y_t | X_t = i \rbr \ ,
\label{eq:obs_like}
\end{eqnarray}  
denote the (conditional) observation likelihoods of the HRC. The observations may
be either discrete or continuous random variables defined on an
appropriate probability space.
 In the continuous observation case, each $C_i(t)$ would denote a probability density function, parametrised by a finite set of parameters. In the discrete (finite) observation case, the $C_i(t)$ would represent a probability mass function with a finite number of mass points. The mixed continuous/discrete observation case is of course, also possible. This observation process models targets in an uncluttered environment, in the following section we present two models of the observation process when there is clutter present.

%% file: single_obs_mod_v8.tex
A general threshold based detection process can be constructed by proposing a model for a detector which has two operating states $\mu_t = \lbr 0, 1 \rbr$ at each time $t$, one for when it detects the target ($\mu_t = 1$), and another for false alarms or non-target detections ($\mu_t = 0$). We denote the \emph{a priori} probabilities over $\mu_t$ as
\begin{align*}
\Pr \lbr \mu_t = 0 \rbr &= \epsilon \\
 \Pr \lbr \mu_t = 1 \rbr &= 1-\epsilon 
\end{align*}\
where $\epsilon \in [0,1]$. The observation model (conditional distribution) for this scenario can then be derived as a mixture of the form
\begin{align}
\Pr \lbr Y_t | X_t = i\rbr & =  \sum_{\mu_t = 0,1} \Pr \lbr Y_t , \mu_t | X_t = i \rbr \nonumber \\
 & =\sum_{\mu_t = 0,1 } \Pr \lbr \mu_t \rbr \Pr \lbr Y_t  | X_t = i, \mu_t \rbr\nonumber\\
 & =  (1-\epsilon) \Pr \lbr Y_t | X_t = i, \mu_t = 1 \rbr \nonumber \\
  & \qquad \qquad \qquad +\epsilon  \Pr \lbr Y_t | \mu_t = 0 \rbr
\label{eq:singl_obs_defn}
\end{align}
We define the first term of Equation \eqref{eq:singl_obs_defn} as another observation likelihood, given we know the operating mode, as $c_i(t) = \Pr \lbr Y_t | X_t = i, \mu_t \rbr$. As before $c_i(t)$ can take any form as in definition \eqref{eq:obs_like}. However, the lower case notation is deliberate as it signifies it is the observation likelihood given the state \emph{and} the operating mode of the sensor.

The final term of \eqref{eq:singl_obs_defn}, represents the likelihood of the observation $Y_t$ given that it is of non-target origin, hence the lack of dependence on the state $X_t$. While the source of such a detection is unknown, it is reasonable to assume it is generated it in a similar way as a target detection and hence can take the same observation likelihood as $c_i(t)$, given we know the operating mode. Specifically, we define
\begin{align*}
\Pr \lbr Y_t | \mu_t = 0\rbr & =  \sum_{j =1}^N \Pr \lbr Y_t  | U_t = j, \mu_t = 0 \rbr \Pr \lbr U_t = j \rbr \nonumber \\
& = \sum_{j =1}^N c_j(t) \Pr \lbr U_t = j \rbr
\end{align*}
where $U_t$ is a temporally uncorrelated process over the statespace representing clutter. For brevity, we will refer to the sensor detector dalse alarm rate $\epsilon$, as the clutter rate.

%% file: filters_v8.tex
Consider a HRC $\Yc$ with state $\Xc$, known MB transition probability matrices  $B_{i,j}^k(t)$,  $t=0,2,\ldots, T-1$,  and known  observation probabilities $C_i(t), t = 0,\ldots, T, i = 1, \ldots, N$. 
Given a sequence of observations, $Y_0,\ldots,Y_T$, the aim here is to compute the filtered {\it a posteriori} probability (APP) mass functions
\begin{align*}
q_{i}(t)  & =  \Pr \lbr X_t = i | Y_1, \ldots, Y_t \rbr \ ,
\end{align*}
for each $t=0,\ldots, T-1$, and each $i=1,\ldots,N$. These APPs can be calculated from the following joint probabilities, for each $k = 1, \dots, N$,
\begin{align*}
q_{i}(t) =  \sum^N_{k=1} \Pr \lbr X_t = i, X_T = k | Y_1, \ldots, Y_t \rbr =  \sum^N_{k=1} q_{i}^{k}(t) \ ,
\end{align*}
via the law of total probability. It has been shown that the joint process $(X_t, X_T)$ is Markov \cite{WC14}, therefore analogously to the hidden Markov model filter (see e.g. \cite{EM02}), it is easily shown via Bayes' rule that $q_i^k(t)$ can be evaluated recursively, for $t=1,\ldots, T-1$ by 
\begin{align}
q_{i}^{k}(t) & = \frac{C_i(t) \, \sum_{j=1}^N B_{j,i}^k(t-1) \, q_{j}^{k}(t-1)}{h(t)} \ ,
\label{eq:std_filt}
\end{align}
where the normalising term is
\begin{align*}
h(t) & =  \Pr \lbr Y_t | Y_0, \ldots, Y_{t-1} \rbr \\
 & = \sum_{i,k=1}^N C_i(t) \, \sum_{j=1}^N B_{j,i}^k(t-1) \, q_{j}^{k}(t-1) \ .
\end{align*}
The overall computational cost of the above filtering recursions is $O(N^3T)$. Initialisation at $t=0$ is via
\begin{align*}
q_{i}^{k}(0) = \frac{C_0(i) \, \Pi_{i,k}}{h(0)},
\end{align*}
\begin{align*}
\mathrm{where}\ \ h(0) = \Pr \lbr Y_0 \rbr = \sum_{i,k=1}^N C_0(i) \, \Pi_{i,k} .
\end{align*}
The observation log-likelihood can be evaluated via $\log \Pr \lbr Y_0, \ldots, Y_T \rbr = \sum_{t=0}^T \log h(t)$. 
This approach avoids any potential numerical underflow issues which might arise if the {\it conditional} (un-normalised) MB filters as defined in \cite{WC11} were used. A track extraction type algorithm can be obtained by comparing the log sequence likelihood to a threshold.